\documentclass[10pt,twocolumn,letterpaper]{article}

\usepackage{cvpr}              

\usepackage[utf8]{inputenc} 
\usepackage[T1]{fontenc}    
\usepackage{url}            
\usepackage{booktabs}       
\usepackage{amsfonts}       
\usepackage{nicefrac}       
\usepackage{microtype}      
\usepackage{xcolor}         
\usepackage{times}
\usepackage{epsfig}
\usepackage{graphicx}
\usepackage{algorithm}
\usepackage{amsmath}
\usepackage{amssymb}
\usepackage{multirow}
\usepackage{arydshln}
\usepackage{wrapfig}
\usepackage{enumitem}
\usepackage{bbding}
\usepackage{float}
\usepackage{textcomp}
\usepackage[font=small]{caption}
\usepackage{tabularx} 
\usepackage{pifont}
\usepackage{listings}
\usepackage{xspace}
\usepackage{gensymb}
\usepackage{makecell}
\usepackage{soul} 
\usepackage{xcolor}  
\usepackage{cuted}
\usepackage{capt-of} 

\definecolor{cvprblue}{rgb}{0.21,0.49,0.74}
\definecolor{urlblue}{rgb}{0.24,0.49,0.9}
\usepackage[pagebackref,breaklinks,colorlinks,urlcolor=urlblue,citecolor=cvprblue]{hyperref}

\makeatletter
\def\adl@drawiv#1#2#3{%
        \hskip.5\tabcolsep
        \xleaders#3{#2.5\@tempdimb #1{1}#2.5\@tempdimb}%
                #2\z@ plus1fil minus1fil\relax
        \hskip.5\tabcolsep}
\newcommand{\cdashlinelr}[1]{%
  \noalign{\vskip\aboverulesep
           \global\let\@dashdrawstore\adl@draw
           \global\let\adl@draw\adl@drawiv}
  \cdashline{#1}
  \noalign{\global\let\adl@draw\@dashdrawstore
           \vskip\belowrulesep}}
\makeatother

\makeatletter
\DeclareRobustCommand\onedot{\futurelet\@let@token\@onedot}
\def\@onedot{\ifx\@let@token.\else.\null\fi\xspace}

\def\eg{\emph{e.g}\onedot}

\def\etal{\emph{et al}\onedot}
\makeatother

\usepackage[capitalize]{cleveref}
\crefname{section}{Sec.}{Secs.}
\Crefname{section}{Section}{Sections}
\Crefname{table}{Table}{Tables}
\crefname{table}{Tab.}{Tabs.}

\definecolor{lightpurple}{RGB}{200, 160, 255}

\definecolor{darkred}{RGB}{210, 0, 0}
\definecolor{darkgreen}{RGB}{0, 200, 50}
\definecolor{darkblue}{RGB}{0, 50, 200}
\definecolor{darkpurple}{RGB}{200, 50, 200}
\definecolor{brightgreen}{rgb}{0.4, 1.0, 0.0}
\definecolor{caribbeangreen}{rgb}{0.0, 0.8, 0.6}

\newcommand\Key[1]{\textcolor{black}{#1}}

\makeatletter
\renewcommand\paragraph{\@startsection{paragraph}{4}{\z@}%
	{0.5ex \@plus.2ex \@minus.2ex}%
	{-1em}%
	{\normalfont\normalsize\bfseries\maybe@addperiod}}
\newcommand{\maybe@addperiod}[1]{#1\@addpunct{.}}
\makeatother

\def\snd{{\mathbf s}}
\def\predfn{f}

\newcommand{\supparxiv}[2]{#1}

\supparxiv{

\def\upvspacefig{\vspace{-2.4mm}}
\def\downvspacefig{\vspace{-2mm}}
\setlength{\belowcaptionskip}{-0mm}
\setlength{\abovecaptionskip}{0.3em}

}{
\def\upvspacefig{\vspace{-0mm}}
\def\downvspacefig{\vspace{-0mm}}
\setlength{\belowcaptionskip}{-0mm}
\setlength{\abovecaptionskip}{0.5em} 

}

\def\paperID{5666} 
\def\confName{CVPR}
\def\confYear{2025}

\title{Supervising Sound Localization by In-the-wild Egomotion}

\author{
Anna Min\textsuperscript{2} \qquad 
Ziyang Chen\textsuperscript{1} \qquad 
Hang Zhao\textsuperscript{2,3} \qquad 
Andrew Owens\textsuperscript{1} 
\vspace{2.5mm}\\
\textsuperscript{1}University of Michigan \quad \textsuperscript{2}Tsinghua University \quad \textsuperscript{3}Shanghai Qi Zhi Institute
}

\begin{document}
\maketitle



\begin{abstract}
We present a method for learning binaural sound localization using egomotion as a supervisory signal. Over the course of a video, the camera's direction to a sound source will change as the camera moves.  We train an audio model to predict sound directions that are consistent with visual estimates of camera motion, which we obtain using traditional methods from multi-view geometry. This provides a weak but plentiful form of supervision that we combine with traditional binaural cues. To evaluate this method, we propose a dataset of real-world audio-visual videos with egomotion. We show that our model can successfully learn from real-world data and that it performs well on sound localization tasks.




\end{abstract}   
\begin{figure}
    \centering
    \includegraphics[width=1.0\columnwidth]{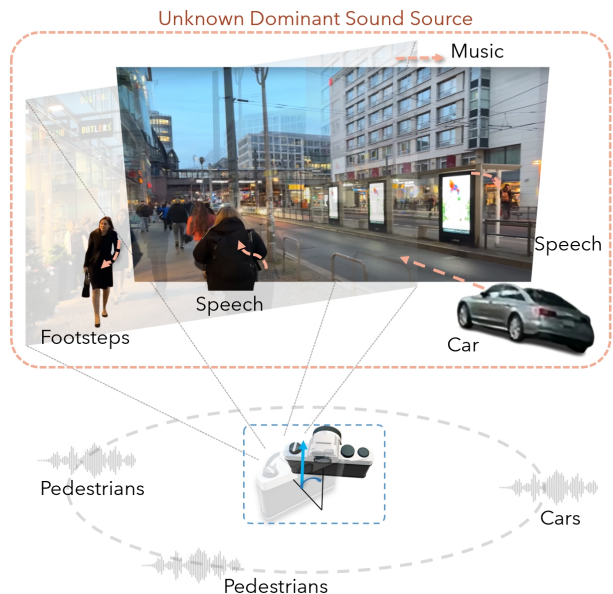}
    \captionof{figure}{{\bf Supervising sound localization using egomotion from natural video.} We use camera motion as a supervisory signal for stereo sound localization. Our model learns to predict changes in sound direction that correlate with changes in visually indicated camera motion. In contrast to prior self-supervised sound localization methods, our model is trained solely on in-the-wild natural videos, which often contain complex camera motions and noisy mixtures of different component sounds. } \label{fig:teaser}  
    \vspace{-2mm}
\end{figure}

\section{Introduction}



Our sense of hearing allows us to perceive events that are out of sight, such as objects that are distant, occluded, or outside of our narrow field of view. Despite the importance of spatial audio perception, existing methods for stereo sound localization often struggle in real-world settings, such as by being limited to synthetic training data~\cite{chen2023sound},  specific binaural cues~\cite{chen2022sound}, or specific visual object categories~\cite{gan2019self}. 

Recent work has addressed this problem using time-delay estimation~\cite{chen2022sound} and jointly learning camera rotation with sound sources~\cite{chen2023sound}. These approaches either rely on training with simulated data or specific hand-chosen binaural cues and thus can only be applied to limited domains. Natural sound recordings vary widely in ways that are difficult to capture through simulation. For example, properties of the equipment itself, such as the geometry of the stereo microphone pair and their frequency responses, are highly varied, and the sounds that they record are often complex mixtures of different sources that make it difficult to extract a learning signal from.

We address the problem of learning stereo sound localization entirely from unlabeled ``in the wild'' videos. We exploit the fact that real-world audio-visual signals contain moments in which the camera moves while the sound source remains approximately stationary, thereby providing a form of cross-modal supervision. 
We use well-established multi-view geometry methods to estimate the relative pose of the camera, and then train a sound localization method to make predictions that are consistent with this motion.

Our technical approach exploits the fact that, while obtaining ground truth information from vision is difficult, {\em approximate} visual information (such as the direction of the camera rotation) are straightforward to estimate and provide constraints that can be used as part of a weakly supervised learning framework. We show that this weak supervision from vision can be combined together with traditional interaural intensity difference cues.

\begin{figure*}[t]
    \centering
    \upvspacefig
\includegraphics[width=1.0\linewidth]{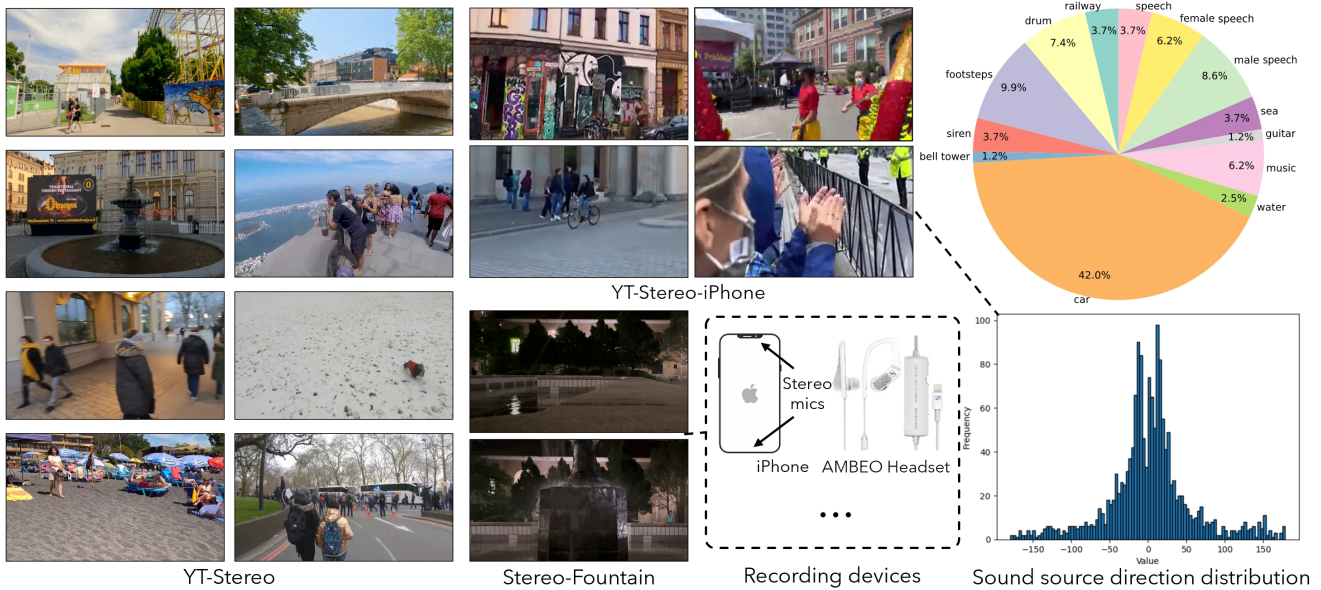}
    \caption{\textbf{{The StereoWalks dataset}.} Example videos from different subsets of our dataset. We focus on in-the-wild audio-visual data sourced from the internet. Our dataset contains YouTube videos recorded with stereo microphones and iPhones, as well as lab-collected in-ear binaural and stereo data.  We also show the estimated distribution of different sound source categories in each subset, to help visualize the contents of each dataset.} 
    \label{fig:dataset_example} 
    \downvspacefig
\end{figure*}

Our method has a number of advantages over prior approaches.  In contrast to other work that uses egomotion~\cite{chen2023sound}, it can be trained entirely on real videos instead of simulated environments. It does not require joint training of visual and audio models; instead, it obtains its learning signal from off-the-shelf multi-view geometry methods. This allows our method to handle diverse motions and camera translation (\cref{fig:teaser}). Moreover, our approach does not require the sound source to be visible~\cite{morgado2018self,gao2019visual-sound}, nor does it place restrictions on the object category~\cite{gan2019self}.


We also propose a dataset containing in-the-wild stereo sound and egomotion and human-provided sound direction labels (the first such dataset, to the best of our knowledge).  Our approach outperforms previous methods trained on simulated data for in-the-wild sound localization. Our contributions are as follows:

\begin{itemize}[leftmargin=10pt, topsep=1pt, noitemsep]
\item We introduce a dataset, called {\em StereoWalks}, for learning sound localization from unlabeled audio-visual signals.
\item We propose a method for training sound localization methods using weak supervision from egomotion.
\item We experimentally evaluate our model in a variety of settings, finding that we obtain better performance than previous methods on our newly proposed sound localization dataset, and obtain performance comparable to other methods on existing real-world benchmarks.

\end{itemize}

\section{Related Work}

\paragraph{Audio-visual spatial perception}
Recent research has explored the spatial correspondence between sight and sound and used them for scene analysis or reasoning ~\cite{chen2021structure,konnoaudio,yang2020telling,morgado2020learning,gao2020visualechoes,yang2022camera,christensen2020batvision,parida2021beyond,zheng2024bat,yun2025spherical,li2025self}. Morgado \etal~\cite{morgado2020learning} use natural audio-visual alignment to extract representations for practical tasks. Gao \etal~\cite{gao2019visual-sound} focus on typical field-of-view videos and binaural audios. Zhou \etal~\cite{zhou2020sep} unify stereo generation and source separation, blending audio-visual features seamlessly. Xu \etal~\cite{xu2021visually} employ spherical harmonic decomposition and head-related impulse response~\cite{algazi2001cipic} to construct pseudo visual-stereo pairs. In recent developments for 3D audio-visual synthesis, Liang \etal~\cite{liang23avnerf} implicitly link audio generation with the 3D geometry and material properties of visual environments, facilitating the creation of immersive videos from various camera perspectives. Chen \etal~\cite{chen2023sound} jointly learn sound source direction and camera pose via self-supervision from multi-view audio-visual data. In contrast, we estimate camera motion using off-the-shelf multi-view geometry methods, and use this as a supervision signal to train an audio model. Consequently, our approach is able to learn from natural video, whereas~\cite{chen2023sound} required simple, simulated scenes as training data.


\paragraph{Spatial audio datasets} 
Spatial audio provides sound direction and distance cues, enriching listeners’ 3D perception. Many researchers have studied these cues to interpret sound in complex scenes. 
Many spatial audio and audio-visual datasets have been collected to address challenges in the Direction of Arrival (DOA) problem~\cite{evers2020locata,guizzo2022l3das22,politis2021dataset,politis2022starss22,adavanne2018sound}.
Shimada \etal~\cite{shimada2024starss23} collects a real audio-visual dataset of spatial recordings of real scenes (rooms) to study sound event localization and detection tasks.
Chen \etal~\cite{chen2024360+} multi-modal panoptic dataset with stereo audio for scene understanding. 
Some researchers also study the sound localization task with synthetic multichannel audio datasets with temporal activation and ground-truth DOA labels~\cite{adavanne2019multi,politis2020dataset,politis2021dataset,guizzo2021l3das21,nagatomo2022wearable,wang2023loss,niu2023experimental}.
Many other audio-visual datasets also contain multi-channel audio~\cite{morgado2018self, morgado2020learning,liang23avnerf, yang2020telling, zheng2023multi, wang2023four} and focus on different tasks. 
However, many of them are constrained by limited hours and sound type~\cite{morgado2018self, gao2019visual-sound} and recorded in lab settings~\cite{gao2019visual-sound,politis2022starss22,shimada2024starss23}. They also require labor-intensive labeling for annotations. 
Chen \etal~\cite{chen2024360+} employs four directional microphones to collect audio signals, while Grauman \etal ~\cite{grauman2024egoexo4dunderstandingskilledhuman} use Aria glasses to record synchronized egocentric audio-visual data.
In contrast to previous datasets, ours is collected from diverse internet videos under an in-the-wild setup, using camera motion signals as free supervision. This approach removes the constraints for specialized hardware like multiple microphones or Aria glasses while capturing more dynamic and chaotic scenarios with complex motion patterns.

\paragraph{Audio-visual learning.} 
Besides spatial correspondence between audio and visual signals, other researchers have explored different topics in audio-visual learning. Some study the deep learning approaches for visual sound localization with audio-visual semantic correspondence~\cite{senocak2018learning,chen2021localizing,zhou2022audio,arandjelovic2017look,owens2018audio}. Some works study the temporal alignment between audio and visual streams~\cite{owens2018audio,feng2023self,chen2021audio,sun2023eventfulness,iashin2024synchformer}.
Owens \etal~\cite{owens2018audio} propose to use self-supervised temporal features for audio-visual scene analysis and apply them to several downstream tasks, \eg, action recognition. 
Many works use visual signals to enhance sound separation~\cite{zhao2019sound, afouras2020selfsupervised, gao2021visualvoice,zhao2018sound}.  
Chen \etal~\cite{chen2024images} explore the visual correspondence between image and spectrogram to create visual spectrograms that that simultaneously look like natural images and sound like natural audio. 
Differing from those works, we focus on using ego motions from vision to supervise sound localization.

\paragraph{Camera motion estimation} Camera motion estimation is an important topic in 3D vision, focusing on accurately tracking camera movement through an environment. Sarlin \etal~\cite{sarlin20superglue} proposed SuperGlue, a method that matches two sets of local features by jointly finding correspondences and rejecting non-matchable points, significantly improving the accuracy of feature matching. In the domain of visual SLAM (Simultaneous Localization and Mapping), Whelan \etal~\cite{whelan2013robust} and Raposo \etal~\cite{raposo2016match} have integrated multiple camera odometry estimation techniques to achieve robust tracking, ensuring that the system can maintain reliable performance even in challenging environments. Furthermore, Rockwell \etal~\cite{rockwell2024far} propose methods for estimating camera pose robustly by combining correspondence-based and learning-based methods.  Our work uses estimated camera poses as supervision for audio, providing a novel approach to enhance audio-visual learning systems.

\section{Method}




When the camera is in motion, the relative movement between sound sources and the camera can provide supervision for sound variations. Building upon this intuition, we use the relative moving direction as supervision, which can be interpreted as the direction of camera rotation direction and translation.
In this section, we first introduce a new spatial audio-visual dataset and then demonstrate how to learn spatial audio information from videos.

\begin{table*}[t]
\newcommand\padd{\phantom{0}}
\centering
\caption{ \textbf{Dataset comparison.} We provide details on dataset length, the proportion of visible sound sources, camera motion types, and sound source types, with each clip representing 5 seconds. Visibility is represented by two numbers: the first indicates the number of sound sources visible for over 4 seconds within the 5-second clip, and the second denotes the number of sound sources visible for less than 4 seconds. IID Binaural acc denotes the accuracy of IID predictions respected by the actual left-right labels. }
\resizebox{\linewidth}{!}{%
\begin{tabular}{llccccccc}
\toprule 
\multirow{2}{*}{Dataset} & \multicolumn{1}{l}{\multirow{2}{*}{Scene subset}} & \multirow{2}{*}{Split} & \multicolumn{2}{c}{Size} & \multirow{2}{*}{\makecell[c]{Visibility\\(\%)}}  &\multicolumn{2}{c}{\small{Motion Type}} &  \multirow{2}{*}{\makecell[c]{IID binaural\\acc~(\%)}} \\
\cmidrule(lr){4-5} \cmidrule(lr){7-8} 
 & & & \small{Clips~(k)} & \small{Duration~(hr)} &  & \small{Camera} & \small{Sound sources} \\
\midrule
\multirow{6}{*}{{\shortstack[c]{StereoWalks\\(ours)}}} & 


\multirow{3}{*}{\makecell[l]{YT-Stereo}}
 &Train & 14.4 & 20.0 & 20 / 50  & Rotation\&Translation  & Unknown & -- \\
&  &Val & \padd 0.1 & \padd  0.2 & 10 / 60 & Mainly Rotation & Moving & 57.5 \\
&  &Test & \padd 0.1 & \padd  0.2 & 10 / 60 & Mainly Rotation & Moving & 57.5 \\



\cdashlinelr{2-9} 
 &
 \makecell[l]{Stereo-Fountain} &Raw & \padd 5.0 & \padd  7.0 & 10 / 30  &Mainly Stationary  & \XSolidBrush &76.2 \\ 
 
\cdashlinelr{2-9} 
& \makecell[l]{Binaural-Fountain} &Raw & \padd 1.4 & \padd 2.0 & 10 / 30  &Mainly Stationary  & \XSolidBrush &98.0  \\


\midrule
L/R Binaural~\cite{chen2022sound} &  \makecell[l]{} & Raw & \padd 1.8 & \padd  3.0 & -- & \XSolidBrush & \XSolidBrush &75.4 \\

\midrule
Simulated & HM3D-SS 2.0 (ours)  & Raw & 21.6 & 30.0 & --  & Rotation\&Translation & Rendered & 97.4 \\
\bottomrule
\end{tabular}
}
\downvspacefig
\label{tab:main_dataset}
\end{table*}

\subsection{The {StereoWalks} dataset}\label{datasets}

Our goal is to learn from in-the-wild audio-visual signals and motion using videos that span a variety of scenes, microphone designs, and cameras. Existing datasets for studying spatial audio are largely recorded in lab settings and primarily contain stationary cameras. To address this, we collect a dataset that we call {\em StereoWalks}. First, we acquire internet video (from YouTube) with camera motions and stereo sound (\eg, walking tours). We obtain a subset that is likely to contain iPhones  (we call the larger set {\em YT-Stereo} and this subset {\em YT-Stereo-iPhone}). To do this, we search for captions and titles mentioning iPhone versions newer than "12S". We filter the raw videos by automatic means such as excluding those with unstable or minimal camera rotation, obtaining 20 hours of video. Human annotators label the validation and test sets with perceived ground truth angles and categorize the dominant sound source. Our dataset ensures audio-visual synchronization since both signals are collected from the same device (the camera with built-in microphones or connected headphones). This synchronization is maintained throughout the data collection and preprocessing pipeline. For further details, please refer to \cref{real-world_datasets} and \supparxiv{the supplementary material}{\cref{appendixA.1}}.

To study the influence of the recording device and additional cues from human ears, we also record two datasets using a commodity stereo microphone (iPhone 13 Pro) and a commodity in-ear binaural microphone (Sennheiser AMBEO Smart Headset). Since the latter is placed inside an ear, the model has access to additional cues due to the influence of the ear shape on the sound. We label these subsets \textit{Stereo-Fountain} and \textit{Binaural-Fountain} respectively. We show some examples in \cref{fig:dataset_example} and the statistics in \cref{tab:main_dataset}.

\subsection{Learning audio localization from egomotion}

We learn to localize sound sources from stereo audio. Following previous work~\cite{chen2023sound,chen2022sound}, we only estimate the sound's azimuth angle, since the other degrees of freedom (elevation and distance) are challenging to perceive from audio alone.

\begin{figure}[t]
    \centering
    \includegraphics[width=\columnwidth]{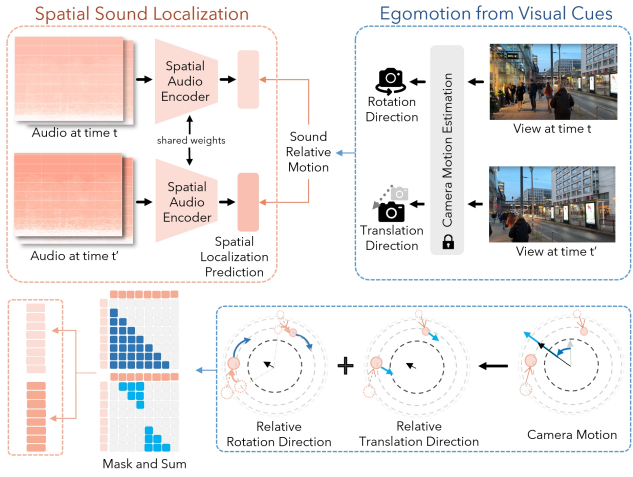}
    \vspace{-3mm}
    \caption{{\bf Method overview.} We use a camera's egomotion to supervise binaural sound localization models. We obtain the rotation and translation of the camera using traditional methods from multi-view geometry. We then train an audio model to predict sound directions that are consistent with visual camera motions, using a dataset that includes in-the-wild walking tours. } \label{fig:method}
    \vspace{-3mm}
\end{figure}



Given an audio clip, we randomly select two short segments, $\snd_1$ and $\snd_2$ from time $t$ and time $t'$ respectively. Since most in-the-wild videos are recorded with stereo audio rather than binaural audio, it remains difficult to predict whether the sound is in front or behind from audio alone if the sound is stationary. We discuss more details in \cref{front-back-confusion} about front-back confusion. We assume all the sound sources are on one side. For each audio clip, we predict an angle $\predfn(\snd_t): \mathbb{R}^{m \times 2} \rightarrow \mathbb{R}^n$, which we represent as a distribution over $n$ angle categories that are uniformly sampled in a grid. The set $A_i$ denotes all the possible angle outputs of $s_t$, where $\forall a \in \ A_i\ ,-90^\circ<a<90^\circ$. 


\paragraph{Camera rotation estimation.} Predicting the direction that a camera is rotating is easier than obtaining its precise angle (\eg, since the latter may require knowledge of camera intrinsics~\cite{hartley2003multiple}). 
We therefore use visual information to predict a binary label $d_r$ that indicates whether the camera is rotating clockwise or counter for a given audio clip. This serves as the pseudo label for training our audio model.


\paragraph{Camera front-back translation estimation.} If the position of the sound source changes slowly compared to the position of the camera, the sound source may appear to move in the opposite direction. We estimate a binary label $d_t$ from the visual information, indicating whether the camera is moving forward for a given audio clip. 
In practice, we use a pre-trained camera pose estimation model to obtain both the translation and rotation labels, $d_t$ and $d_r$.

\paragraph{Supervising audio localization using sight.}

The visually estimated camera rotation estimates $d_r$ restrict the pairs of angles that $\snd_1$ and $\snd_2$ can compatibly be assigned to. We introduce a loss: 
\begin{equation}
  \mathcal{L}_{\mathtt{rot}} = L_{\mathtt{ce}}\left(\sum_{(i, j) \in R} \predfn(\snd_1)_i \predfn(\snd_2)_j, d_r\right),
\end{equation}
where $L_{\mathtt{ce}}$ is cross entropy loss and $R = \{(\alpha,\beta)~:~\alpha \in A_i, \beta \in A_j, \sin(\beta-\alpha) \cdot d_r > 0\}$ is the set of angle labels that are compatible with. 

Similarly, the visually estimated camera front-back translation estimates $d_t$ restrict the pairs of angles that $\snd_1$ and $\snd_2$ can compatibly be assigned to the audio. We introduce a loss:
\begin{equation}
  \mathcal{L}_{\mathtt{trans}} = L_{\mathtt{ce}}\left(\sum_{(i, j) \in T} \predfn(\snd_1)_i \predfn(\snd_2)_j, d_t\right),
\end{equation}
where $L_{\mathtt{ce}}$ is cross entropy loss and $T = \{(\alpha,\beta)~:~\alpha \in A_i, \beta \in A_j, (|\beta|-|\alpha|) \cdot d_t > 0\}$ is the set of angle labels that are compatible with the visual prediction. To prevent the model from collapsing into a trivial solution (for example, predicting all angles to be zero), we exclude cases where the two angles are equal.

\paragraph{Binaural cues} Interaural intensity difference (IID) cues are commonly used in learning-based sound localization works. We use these cues within our model to provide a complementary form of supervision. We follow ~\cite{chen2023sound} and use the difference in loudness between the left and right channels to determine which side the sound is on. We penalize predictions that are inconsistent with these predictions:
\begin{equation}
    \mathcal{L}_{\mathtt{bin}} = L_{\mathtt{ce}}\left(\sum_{j \in B}f(\snd_i)_j, b_t\right),
\end{equation}
where $B$ is the set of angle labels that is consistent with the label $b_t$, which indicates whether the sound is on the left or right side of the camera. We average this loss over the pair of audio clips $\snd_1$ and $\snd_2$.

\paragraph{Overall loss.} Combining these losses together, we minimize:
\vspace{-2mm}
\begin{equation}\mathcal{L} =   \lambda_1 L_{\mathtt{rot}} + (1-\lambda_1) L_{\mathtt{trans}} + \lambda_2  L_{\mathtt{bin}} ,
\end{equation}
where  $\lambda_1$ and $\lambda_2$ control the relative weight of the two losses. The losses are essentially the sum of individual cross-entropy losses between the prediction and each ``allowed angle'' after applying the constraints. As shown in \cref{fig:method}, ``mask and sum'' denotes summing over only the valid labels.

\begin{table*}[t]
\upvspacefig
  \centering
  \caption{Comparison with state-of-the-art methods and other self-supervised methods on YT-Stereo and In-the-wild dataset. \textit{Sim.} denote the simulated dataset.}
  \label{fig:main_table}

\begin{minipage}[t]{0.78\linewidth}
\centering
  \resizebox{1.0\columnwidth}{!}{
  \setlength{\tabcolsep}{2pt}
    \begin{tabular}{lccccccccccc}
    \toprule 
    \multirow{3}{*}{Model} & \multicolumn{2}{c}{Training Set} & \multicolumn{9}{c}{Test Set} \\
    \cmidrule(lr){2-3} \cmidrule(lr){4-12}  
    & \multirow{2}{*}{\shortstack{YT-Stereo}} & \multirow{2}{*}{Sim.} & \multicolumn{3}{c}{YT-Stereo-iPhone} & \multicolumn{3}{c}{Stereo-Fountain} & \multicolumn{3}{c}{Simulated}\\
    \cmidrule(lr){4-6}  \cmidrule(lr){7-9}  \cmidrule(lr){10-12} 
    &  &  & MAE~(\degree)$\downarrow$ & 2clf~(\%)$\uparrow$ & 8clf~(\%)$\uparrow$ & MAE~(\degree)$\downarrow$ & 2clf~(\%)$\uparrow$ & 8clf~(\%)$\uparrow$  & MAE~(\degree)$\downarrow$ & 2clf~(\%)$\uparrow$ & 8clf~(\%)$\uparrow$ \\
    \midrule
    Chance & & & 55.3 & 46.7 & 12.7 & 62.1 & 52.0 & 16.0 & 39.4 & 49.0 & 15.0 \\
    IID -- direct & &  & -- & 57.5 & -- & -- & 97.0 & -- & -- & 97.4 & -- \\
    GTRot~\cite{chen2023sound} & \checkmark &  & 71.4 & 54.0 & 7.1 & 57.2 & 97.0 & 18.4 & 40.2 & 48.7 & 18.7 \\
    \cdashlinelr{1-12} 
    Ours -- IID only~\cite{chen2023sound} & \checkmark & & 37.3 & 54.7 & 28.0 & 30.1 & 97.0 & \textbf{46.0} & 88.3 & 48.7 & 6.8 \\
    Ours -- Simulated & & \checkmark  & 73.4 & 61.7 & 13.3 & \textbf{28.5} & 85.7 & 22.3 & \textbf{9.8} & \textbf{98.3} & \textbf{63.4} \\
    \cdashlinelr{1-12} 
    Ours -- Full & \checkmark & & \textbf{34.0} & \textbf{61.7} & \textbf{33.3} & 29.3 & \textbf{97.3} & \textbf{46.0} & 57.2 & 48.7 & 10.7 \\
    \bottomrule
    \end{tabular}
  }
  \end{minipage}
\hfill
\begin{minipage}[b]{0.21\linewidth}
  \centering
  \resizebox{1.0\columnwidth}{!}{
  \setlength{\tabcolsep}{2pt}
    \begin{tabular}{lc}
    \toprule
    In-the-wild audio~\cite{chen2022sound} & 2clf~(\%)$\uparrow$ \\
    \midrule
    Chance & 50.0 \\
    IID -- direct & 75.4 \\
    GCC-PHAT & 77.2 \\
    GTRot~\cite{chen2023sound} & 84.9 \\
    MonoCLR~\cite{chen2022sound} & 87.4 \\
    StereoCRW~\cite{chen2022sound} & 87.2 \\
    \midrule
    Ours -- IIDonly & \textbf{87.7} \\
    \cdashlinelr{1-2} 
    Ours -- full & 87.5 \\
    \bottomrule
    \end{tabular}
 }
  \end{minipage}
\end{table*}

\section{Experiments}

\subsection{Datasets}
\paragraph{Real-world dataset}\label{real-world_datasets}
To evaluate our approach with the real-world data, we label a subset of \textit{YT-Stereo-iPhone} and \textit{Stereo-Fountain} as mentioned in \cref{datasets} and \cref{tab:main_dataset}. 
To ensure stable sound sources for the validation and test sets, we filter the dataset and label the audio sources through several steps. First, we segment each video into 5-second clips with a 3-second overlap. Then, we estimate the horizontal movement (left/right) using the SuperGlue model and calculate Interaural Intensity Difference (IID) cues. We select videos with large changes in IID. For initial filtering, we compute an IID score every second by calculating the IID for the preceding 2 seconds, sorting the absolute product of IID changes in descending order. Videos are then discarded if the camera rotation direction differs from the IID change direction or if the rotation angle is minimal. After filtering, we label each clip with the location of the audio sources and classify the audio type (such as car, male/female speech, sea, or animals).

\paragraph{Simulated dataset}
To study how visual cues supervise sound localization, we created a simulated dataset to supplement our in-the-wild data. This dataset contains ground truth sound source locations and egomotion. Following~\citet{chen2023sound}, we used the SoundSpaces 2.0~\cite{chen22soundspaces2} platform to construct a dataset, denoted \textit{HM3D-SS 2.0}, that incorporates more diverse settings of camera motion (rotation and translation) and sound source motion. Specifically, we simulate some moving audio source and include translation in the egomotion as well. We generate the dataset with binaural Room Impulse Responses, using 3D scenes from the Habitat-Matterport 3D~\cite{ramakrishnan2021habitat}. We partitioned the data into train/validation/test sets based on scenes. To create binaural audio, we convolve Room Impulse Responses (RIRs) with mono-channel audio from LibriSpeech~\cite{panayotov2015librispeech}. To align with in-the-wild data, we constrained the total rendered audio length to about 30 hours, matching the in-the-wild dataset’s scale. The \textit{HM3D-SS} dataset from \citet{chen2023sound} is a subset of our \textit{HM3D-SS 2.0} dataset, as shown in Setting (1) of \cref{tab:overlap_sound}. Please see \supparxiv{the supplementary material}{\cref{appendixA.1}} for details. 

\subsection{Implementation details} 
For the angle prediction model, we used a ResNet-18~\cite{he2016identity} architecture on spectrograms. We transformed the two-channel waveform of length $L$ into a $256 \times 256 \times 4$ spectrogram using a short-time Fourier transform, retaining both magnitude and phase. We extracted features from the spectrograms and mapped them to 1-dimensional logits for $n$-class classification, setting $n=32$. For camera motion, we used a pre-trained Superglue model. The training was conducted on an 80GB A100 GPU with a batch size of 300 for 100 epochs. For Ours-full, $\lambda_1$ was set to 0.9 and $\lambda_2$ to 1; for Ours-R\&T, $\lambda_1$ was 0.9 and $\lambda_2$ was 0. In \cref{fig:main_table}, we first train on \textit{YT-Stereo} and part of \textit{YT-Stereo-iPhone}, then fine-tune for one epoch on each dataset.

To obtain the camera motion, we employ the Superglue model~\cite{sarlin2020superglue} and utilize the perspective field~\cite{jin2022PerspectiveFields} to predict the horizontal field of view, resulting in the rotation matrix and the translation vector. We sampled 5 frames per second for a 5 sec. video and calculated the rotation matrix and translation vector between each frame. Additionally, we computed the rotation matrix and translation vector between different images at intervals of 3 and 6 frames. For the purposes of cleaning the dataset, for any two time points, we accumulated the rotations calculated at intervals of 3 frames, proportionally adjusting if there were gaps. The same approach was applied to translations.

\begin{table*}[t]
  \centering
  \small
  \setlength\tabcolsep{12pt}
  \caption{360-Degree Sound Localization Prediction Results for YT-Stereo-iPhone and Stereo-Fountain Datasets. The YT-Stereo-iPhone dataset includes rotations and translations with mostly moving sound sources, while the Stereo-Fountain dataset has mostly stationary cameras and sound sources.}
  
  \resizebox{0.9\linewidth}{!}{
  \begin{tabular}{@{}lcccccccc@{}}
    \toprule
    \multirow{2}{*}{Model} & \multicolumn{3}{c}{YT-Stereo-iPhone} & & \multicolumn{3}{c}{Stereo-Fountain} \\
    \cmidrule(lr){2-4} \cmidrule(lr){6-8}
    & MAE~(\degree)~$\downarrow$ & 2clf~(\%)~$\uparrow$ & 8clf~(\%)~$\uparrow$ & & MAE~(\degree)~$\downarrow$ & 2clf~(\%)~$\uparrow$ & 8clf~(\%)~$\uparrow$ \\
    \midrule
    Chance & 55.3 & 46.7 & 12.7 & & 62.1 & 52.0 & 16.0 \\
    IID-direct & -- & 57.5 & -- & & -- & 97.0 & -- \\
    GTRot~\cite{chen2023sound} & 71.4 & 54.0 & 7.1 & & 55.1 & 97.1 & 18.7 \\
    Ours -- Simulated & 73.4 & \textbf{61.7} & 13.3 & & 28.5 & 85.7 & 22.3 \\
    \cdashlinelr{1-8}
    Ours -- IID only~\cite{chen2023sound} & 37.3 & 54.7 & 28.0 & & 29.6 & 97.3 & 46.0 \\
    Ours -- R\&B & 34.5 & 55.4 & \textbf{33.4} & & 28.5 & 97.3 & 46.0 \\
    Ours -- T\&B & 37.4 & 53.4 & 26.8 & & -- & -- & -- \\
    Ours -- Full & \textbf{34.0} & \textbf{61.7} & 31.7 & & \textbf{28.5} & \textbf{97.3} & \textbf{46.0} \\
    \midrule
    Supervised & -- & -- & -- & & 33.8 & 98.0 & 49.5 \\
    Supervised -- RTF & -- & -- & -- & & 27.1 & 98.0 & 49.5 \\
    \bottomrule
  \end{tabular}
  }
  \downvspacefig
  \label{tab:main_other_datasets}
\end{table*}

\subsection{Evaluation}

\paragraph{Baselines and ablations}


We evaluate the performance of sound localization by the MAE of angle prediction and the accuracy of 2-way classification to compare with IID-direct and 8 classification. The accuracy of 2 classification is denoted as ``2clf'' and that of 8 classification is denoted as ``8clf''. Since there exists front-back confusion for stereo devices if given only one audio segment, we reflect all the predictions to the front to avoid ambiguity.

We use the following baselines: 1) Chance: Baseline performance by random guessing. 2) IID-direct: Direct prediction using interaural intensity difference (IID) cues, as described in Chen et al.~\cite{chen2023sound}. 3) GTRot: Implementation of the method from Chen et al.~\cite{chen2023sound}, which uses ``oracle'' rotation angle and pseudo binaural cues as supervision. For real-world datasets, we implemented GTRot training using pseudo labels of rotation angles obtained by the SuperGlue model~\cite{sarlin2020superglue} and Perspective Fields~\cite{jin2022PerspectiveFields} for the \textit{YT-Stereo-iPhone} dataset. For the Stereo-Fountain dataset, the angles are labeled by humans for the training set and evaluation set. 4) Supervised: We train a supervised model using ground-truth angles for each dataset, covering 360 degrees. 5) Supervised--RTF: We train a supervised model using ground truth mapping of the sound source behind the recording device to the front, followed by supervised learning.

For evaluating on the in-the-wild binaural audio benchmark~\cite{chen2022sound}, we use binary classification accuracy as the metric. Apart from the baselines above, we compare our model to a state-of-the-art self-supervised binaural prediction method based on time delay prediction and contrastive random walks trained on stereo sounds~\cite{chen2022sound}. MonoCLR denotes the model with instance discrimination trained on mono sounds. GTRot~\cite{chen2022sound} is trained on simulated data HM3D-SS from SoundSpaces 2.0~\cite{chen22soundspaces2}. StereoCRW and MonoCLR are trained with FMA music samples~\cite{defferrard2016fma} and FAIR-Play~\cite{gao2019visual-sound}. The IID-direct is trained with the In-the-wild audio of ~\citet{chen2022sound}.  

We investigated several model variants to determine whether our model effectively uses visual cues. 1) \textbf{Ours--full}: using $L_{rot}$, $L_{trans}$ and $L_{b}$, 2) \textbf{Ours--IIDonly}: employing $L_{b}$ only, 3) \textbf{Ours--R\&B}: using $L_{rot}$ and $L_{b}$ without $L_{trans}$, 4) \textbf{Ours--T\&B}: using $L_{trans}$ and $L_{b}$ without the rotation loss, 5) \textbf{Ours--R\&T}: using $L_{rot}$ and $L_{trans}$ without the pseudo binaural loss, which is only supervised by the egomotion labels, 5) \textbf{Ours--simulated}: training models solely on the simulated dataset \textit{HM3D-SS 2.0} Dataset with three losses.

In GTRot, Supervised, Supervised–RTF, and our models, the output has 32 classification possibilities. We calculate the Mean Absolute Error by summing the product of each possibility and the midpoint of its corresponding interval.

\subsection{Results}

\begin{figure}
    \centering
    \includegraphics[width=\linewidth]{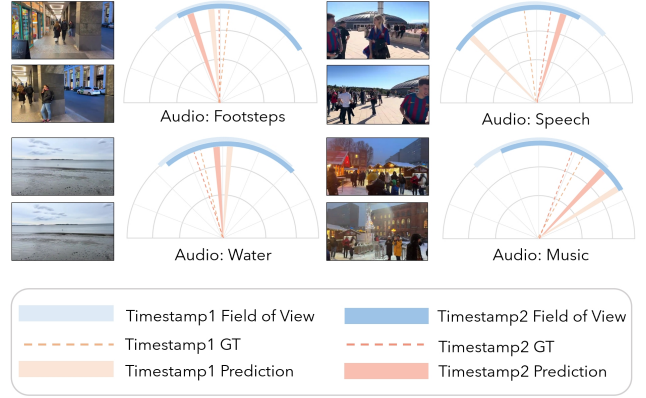}
    \vspace{-0.3cm}
    \caption{Visualizations of results in the internet walking tour videos (the \textit{YT-Stereo} subset). We show our predictions and ground-truth annotations with angles and audio class labels. 
    }
    \label{fig:visualization_results}
    \vspace{-0.5cm}
\end{figure}

As shown in \cref{fig:main_table}, our proposed method, Ours-full, outperforms baselines across various in-the-wild datasets. Tested on the \textit{YT-Stereo-iPhone} dataset, our model achieved the best results, showing advantages over other baselines. We tested our models on both \textit{YT-Stereo-iPhone} and \textit{Stereo-Fountain} datasets, demonstrating robust performance across scenarios. Visualizations are shown in \cref{fig:visualization_results}. Our model outperforms the method in \cite{chen2023sound}, which uses stronger supervision.

When considering the sim2real gap, our simulated model has not generalized well to \textit{YT-Stereo-iPhone} and \textit{Stereo-Fountain}. The model trained on real-world data does not generalize well to the simulated dataset. This highlights the necessity of training with ego-motion as supervision on in-the-wild data for spatial sound localization.

As shown in \cref{tab:main_other_datasets}, when training on specific datasets, Ours-full achieves comparable best performance on \textit{YT-Stereo-iPhone} and \textit{Stereo-Fountain}, demonstrating the effectiveness of our method. Evaluating on the previous dataset In-the-wild audio, we fine-tuned our model trained on \textit{YT-Stereo}, and it achieved comparable good results.

\begin{table*}
\upvspacefig
\centering
\captionof{table}{{Overlapping sounds make in-the-wild sound localization challenging: Experiments on three simulated datasets where the camera makes small translations in all settings. To better align with the real-world experiments, as in settings (1)(2)(3), the dataset is restricted to 30 hours, while in (4), the dataset is not limited. ``Intermittent'' refers to sound sources that are silent for half the time. ``Overlap'' refers to the presence of multiple sound sources.} }
\resizebox{\linewidth}{!}{
\begin{tabular}{l ccc ccc ccc ccc}
    \toprule
    \multirow{2}{*}{Model}      
     & \multicolumn{3}{c}{(1) One Source} & \multicolumn{3}{c}{(2) Overlap } & \multicolumn{3}{c}{(3) Overlap Intermittent} &\multicolumn{3}{c}{(4) More Simulated Data of (3)}\\
      \cmidrule(lr){2-4}  \cmidrule(lr){5-7}  \cmidrule(lr){8-10} \cmidrule(lr){11-13}& MAE(\degree)$\downarrow$ & 2clf(\%)$\uparrow$ & 8clf(\%)$\uparrow$ & MAE(\degree)$\downarrow$ &2clf(\%)$\uparrow$ &8clf(\%)$\uparrow$  & MAE(\degree)$\downarrow$ &2clf(\%)$\uparrow$ &8clf(\%)$\uparrow$ & MAE(\degree)$\downarrow$ &2clf(\%)$\uparrow$ &8clf(\%)$\uparrow$\\
    \midrule
    Chance & 39.4 &49.0 &15.0 &39.4 &49.0 &15.0 & 39.4 &49.0 &15.0 & 39.4 &49.0 &15.0 \\
    IID &  - &97.4 &- &  - &80.1 &- &  - &80.8 &- &  - &80.8 &- \\
    GTRot~\cite{chen2023sound} &  \textbf{ 4.3 }&97.5 &\textbf{84.7 }&\textbf{22.5 }&83.5&33.5 &26.9 & 81.5 & 32.5 &\textbf{19.7} & \textbf{87.2} & 41.0 \\
    \cdashlinelr{1-13} 
    Ours -- IIDonly~\cite{chen2023sound} & 22.2 &\textbf{99.0} &35.7 & 25.6 &78.8 &26.1 & 28.0 & 78.0 & 23.7 & 26.7 & 79.5 & 25.7\\
    \cdashlinelr{1-13} 
    Ours -- Full &  9.8 &98.3 &63.4 & 23.2 &\textbf{87.2} &\textbf{36.2} &\textbf{24.1} & \textbf{82.1} & \textbf{35.7} & 23.7 & 82.4 & \textbf{41.2}\\
    \midrule
    Supervised   &2.8 &98.7&89.8 & 10.8 & 91.8 & 70.2 & 8.4 & 94.1 & 74.6 & 5.7 & 96.0 & 80.2 \\
    \bottomrule
\end{tabular}
}
\label{tab:overlap_sound}

\end{table*}

\begin{table*}[ht]
\centering
\caption{Relationship between ego-translation and relative motion of sources. Experiments were conducted with various types of motion and distances from the camera using the simulated dataset \textit{HMSS-3D 2.0}  made by SoundSpaces 2.0~\cite{chen22soundspaces2}. We report Mean Absolute Error in the unit of degrees (\degree), and accuracy of 2clf and 8clf in the unit of percentage (\%).}
\resizebox{\linewidth}{!}{
\begin{tabular}{l ccc ccc ccc ccc ccc ccc}
    \toprule
    \multirow{2}{*}{Model}      
     & \multicolumn{3}{c}{(1) Rotation Only} & \multicolumn{3}{c}{(2) Translation Only} & \multicolumn{3}{c}{(3) Rotation\&Translation} & \multicolumn{3}{c}{(4) Distant Rotation} & \multicolumn{3}{c}{(5) Distant R\&T} & \multicolumn{3}{c}{(6) Overlap R\&T} \\
      \cmidrule(lr){2-4}  \cmidrule(lr){5-7}  \cmidrule(lr){8-10}  \cmidrule(lr){11-13} \cmidrule(lr){14-16} \cmidrule(lr){17-19} & MAE$\downarrow$ & 2clf$\uparrow$ & 8clf$\uparrow$ & MAE$\downarrow$ &2clf$\uparrow$ & 8clf$\uparrow$  & MAE$\downarrow$ &2clf$\uparrow$ &8clf$\uparrow$  & MAE$\downarrow$ &2clf$\uparrow$ &8clf$\uparrow$  & MAE$\downarrow$ &2clf$\uparrow$ &8clf$\uparrow$ & MAE$\downarrow$ &2clf$\uparrow$ &8clf$\uparrow$  \\
    \midrule
    Chance & 39.4 &49.0 &15.0 &39.4 &49.0 &15.0 &39.4 &49.0 &15.0 &39.9 &49.2 &15.7 &39.8 &49.1 &15.8 &39.0 &49.5 &15.2\\
    IID &  - &97.4 &- &  -  &97.4 &-  &  -  &97.4 &-   & - & 95.6 & - &  -  &95.1 &- &  -  &73.5 &-  \\
    
    GTRot~\cite{chen2023sound} &  \textbf{ 4.3} &97.5 &\textbf{84.7} & -  & - & -  & 4.4 & 97.5 & 84.5 & \textbf{4.5} & 97.1 & \textbf{82.0 }& \textbf{4.7} & 96.7 & \textbf{86.9} & 28.1 & 79.6 & 30.1\\
    \cdashlinelr{1-19} 
    Ours -- IIDonly~\cite{chen2023sound} & 22.2 &\textbf{99.0} &35.7 &\textbf{22.7}&\textbf{99.0} &\textbf{35.7} & 22.0 & \textbf{99.1 }& 35.9  & 23.1 & 97.8 & 35.2 & 23.0 & 98.1 & 35.0 & 32.5 & 75.0 & 25.7 \\
        Ours -- R\&B &  9.8 &98.3 &63.4 & -  & - & - & \textbf{10.1 }& 98.0 & \textbf{60.1} & 10.7 & \textbf{98.1} & 59.1 & 11.7 & \textbf{98.6} & 58.7 & \textbf{25.6} & \textbf{81.4} & \textbf{34.6} \\
    Ours -- T\&B & -  & - & - &23.2 &96.0 &\textbf{35.7} &23.0 &95.7 &35.0 & - & - & - & 22.7& 96.7&35.1  & 30.1 & 72.5 & 22.3\\
    Ours -- R\&T & -  & - & - & -  & - & - & 14.7 &96.3 &48.3  & - & - & -  & 22.5 & 95.0 & 45.7 & 26.7 & 79.0 & 29.2\\
    \cdashlinelr{1-19} 
    Ours -- Full &  9.8 &98.3 &63.4 &23.2 &96.0 &\textbf{35.7 }& 11.7 & 98.0 & 58.7  & 11.2 & \textbf{98.1} & 56.8 & 11.9 & 97.6 & 55.3 & 28.6 & 76.0 & 28.5 \\
    \midrule
    Supervised  & 2.8 &98.7 &89.6 & 2.6 &99.1 &89.9 & 2.6 &98.9 &89.5  & 2.7 & 98.8 & 90.0 & 2.5 & 99.0 & 90.4 & 12.1 & 86.0 & 65.4 \\
    \bottomrule
\end{tabular}
}
\downvspacefig
\label{fig:ablation_motion}
\end{table*}

\subsection{Analysis}

\paragraph{Overlapping sound assumption} 

Previously, based on the hypothesis that co-occurring audio and visual signals offer ``free'' supervision capable of capturing geometry, including camera motion and sound source direction, Chen \etal~\cite{chen2023sound} proposed jointly estimating camera rotation from images and sound direction from binaural audio. Their approach leverages geometric consistency and pseudo-binaural cues. In that work, audio events are mostly singular or the main sound source is significantly louder than secondary sources, making the sound environment relatively ideal. However, in real-world scenes, sounds may overlap and be intermittent, making  localization more challenging.

We assume that the overlapping and intermittent sounds contribute to our model's better generalization compared to GTRot, the ``oracle'' method from~\cite{chen2023sound}, despite GTRot having stronger supervision than our model. To test these hypotheses, we conducted experiments in a simulated environment based on SoundSpaces 2.0~\cite{chen22soundspaces2}. Since most sounds in natural video are overlapping and intermittent sounds, we tested different models under the following settings: (1)~{\em One Source}: Only one sound source in the environment. (2)~{\em Overlap}: A secondary sound source is added at a constant position, with a loudness of 0.7 relative to the main sound source. (3)~{\em Overlap Intermittent}: A secondary sound source is added with 0.7 times the loudness of the intermittent main sound source, where both sources are silent independently for half of the time. (4)~{\em More Simulated Data of (3)}: We generate additional data by pairing room impulse responses with mono-channel audio for convolution, creating more data pairs. In these settings, camera translations are controlled within a small range, where the pose is restricted to a small area. To better align with the real-world experiments, as in settings (1)(2)(3), the dataset is restricted to 30 hours, while in (4), the dataset is not limited.
 

As shown in \cref{tab:overlap_sound}, with limited data, our proposed method surpasses previous methods in settings (2)~{\em Overlap} and settings (3)~{\em Overlap Intermittent} while in settings (1)~{\em One Source} GTRot has the best performance. This outcome suggests that using egomotion direction instead of angle as supervision can generalize better to more complicated scenarios. Regarding settings (4)~{\em More Simulated Data of (3)}, although GTRot has stronger supervision than Ours-full, they show comparable performance, demonstrating our method’s effectiveness.

\paragraph{Rotation and translation} 
In the internet walking tour videos (the \textit{YT-Stereo} subset), since most sounds are distant, the absolute translation of the sound sources greatly affects the relative translation between the sound source and the camera. However, in terms of angular velocity, the rotation speed of the camera itself is much greater than the rotational speed of the sound source relative to the camera. Therefore, we can disregard the relative translation. From experimental results, it can be observed that removing the translation loss from the model yields performance on par with models that include translation loss on the \textit{YT-Stereo} dataset.

To study how camera rotation and translation affect sound localization and identify which loss functions are dominant in different scenarios, we conducted experiments in a simulated environment with the following settings as shown in \cref{fig:ablation_motion}: (1)~{\em Rotation Only}: Only camera rotation with a single sound source. (2)~{\em Translation Only}: Only camera translation with front-back translation in the range of 0--2m. (3)~{\em Rotation\&Translation}: Both camera rotation and front-back translation, with translation in the range of 0--2m. (4)~{\em Distant Rotation}: A sound source 5 meters away from the camera, otherwise identical to setting (1). (5)~{\em Distant R\&T}: A sound source 5 meters away from the camera and within 7 meters, otherwise identical to setting (3). (6)~{\em Overlap R\&T}: Two sound sources, with the addition of a secondary sound source with a constant position and 0.7 loudness to the main sound source, otherwise identical to setting (3)~{\em Rotation\&Translation}. In all settings, the dominant sound sources have their movement within 0.5 meters.


Without relying on pseudo-binaural cues, our method—guided solely by camera ego-motion—achieves similar performance in setting (3)~{\em Rotation\&Translation}, where the camera translates within 0–2 meters. This suggests that for nearby sound sources, small camera translations provide sufficient cues for sound localization. However, in setting (5)~{\em Distant R\&T}, a new challenge arises as sound sources are positioned further away, making camera shifts less impactful on sound positions and resulting in minimal angle changes.

Models trained with both rotation and pseudo-binaural cues effectively capture distant sound localization cues. In setting (6)~{\em Overlap R\&T}, with overlapping sound sources and complex motion, our ego-motion-based approach continues to perform comparably to the best results. This highlights the method’s adaptability, using camera movements to accurately localize sounds across various challenging conditions.





\begin{table}[t]
\centering
\caption{Evaluation of sound localization on the \textit{Stereo-Fountain} and \textit{Binaural-Fountain} datasets, showing improved accuracy in front/back localization with binaural recordings.}
\resizebox{\linewidth}{!}{%
\begin{tabular}{lccc}
\toprule
Dataset & MAE~$\downarrow$ & Left/Right Acc~$\uparrow$ & Front/Back Acc~$\uparrow$ \\
\midrule
Stereo-Fountain & 33.8 & 98.0\% & 51.0\% \\
Binaural-Fountain & \textbf{27.8} & \textbf{99.0\%} & \textbf{69.3\%} \\
\bottomrule
\end{tabular}%
}

\label{tab:frontback_confusion_results}
\end{table}


\paragraph{Front-Back ambiguity}\label{front-back-confusion}


To explore how different recording devices handle front-back sound localization, we conducted an experiment to predict supervised $360^\circ$ sound localization on \textit{Stereo-Fountain} and \textit{Binaural-Fountain}, both recorded beside the same fountain using the microphones of an iPhone 13 Pro and a Sennheiser AMBEO Smart Headset, respectively. The recordings allowed us to compare how well each setup managed $360^\circ$ sound localization.


As shown in \cref{tab:frontback_confusion_results}, the model obtained higher accuracy on \textit{Binaural-Fountain} than \textit{Stereo-Fountain}. 
The latter dataset was recorded on an iPhone. Its stereo microphone may have less variation in the audio captured from behind versus in front of the device. In contrast, the in-ear Sennheiser AMBEO Smart Headset offered a clearer distinction, which may be due to the extra cues provided by the wearer's ear or head. 

\section{Conclusion}

We propose a method and dataset for in-the-wild sound localization. Our method is trained without labeled data. Instead, it obtains its supervision from visual camera motion and binaural audio cues. Through our experiments on in-the-wild and simulated data, we find that this egomotion supervision aids sound localization. 

\paragraph{Limitations} 
While our work introduces the first in-the-wild sound localization training dataset and corresponding evaluation sets, the challenges of low-quality stereo audio and prevalent overlapping sounds complicate our data-cleaning process. Filtering valid clips from numerous unlabeled monocular videos captured in the wild may result in overlooking brief sound effects.


\paragraph{Future work} We anticipate that future research will address these limitations by incorporating more visual and contextual data. We see our work as a step toward creating sound localization methods that work in challenging real-world conditions, and a step toward multimodal methods that learn geometric information from unlabeled data.  

\paragraph{Acknowledgements} This work was completed as part of an internship by Anna Min at University of Michigan. We would like to thank Zafar Rafii, Ethan Manilow, and Ruohan Gao for their helpful feedback. We thank John Chu for providing samples he recorded on YouTube for preliminary experiments.

{
    \small
    \bibliographystyle{ieeenat_fullname}
    \bibliography{main}
}

\newpage
\clearpage
\setcounter{page}{1}
\maketitlesupplementary

\renewcommand{\thesection}{A.\arabic{section}}
\setcounter{section}{0}

\section{Dataset}
\label{appendixA.1}

We provide more information about our dataset collection procedure. 

\subsection{Internet data collection}
One of the major components of our dataset is the {\em YT-stereo} subset, which is obtained by querying YouTube for queries associated with walking tours, such as ``walking tour."  We filter out videos shot in portrait mode or with mono audio. 

\Key{To ensure that the camera's motion correlates with the relative change of the sound sources, it's essential that the angular velocity of the sound source is much slower than the camera's angular velocity.} \Key{To get stable sound sources as the validation set and test set, which provide a stronger supervision signal, we take a way to filter the dataset for stable sound sources.} After the process, 20 hours of videos are selected for the training set.

\begin{figure}[ht]
    \centering
    \includegraphics[width=\linewidth]{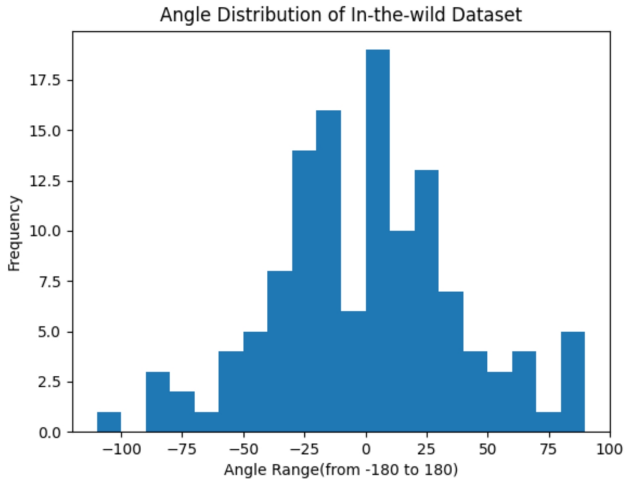}
    \caption{The angle distribution of YT-Stereo-iPhone validation set. The angles range from -180 to 180 without filtering.} \label{fig:distribution}
\end{figure} 

\

\subsection{Dataset Filtering}

\paragraph{YT-Stereo-iPhone} 
The training set is unfiltered, while the validation and test sets are filtered using the following methods.

We manually select 100 clips as the validation set from a pool of 1600 filtered clips.
To ensure that the motion of the camera correlates with the relative change of the sound sources, it is essential that the angular velocity of the sound source is much slower than the angular velocity of the camera. To get clips of clear camera motion for training, we filter the dataset with the following steps: 1) We segment the videos into 5-second clips every 3 seconds. 2) We estimate horizontal motion using the Superglue model~\cite{sarlin20superglue} to get the rotation matrix and translation matrix of the clips. 3) We filter out videos with camera rotation angles of less than 10 degrees.


\begin{table*}[t]
\centering
\caption{ \textbf{Dataset comparison.} We provide details on dataset length, the proportion of visible sound sources, camera motion types, and sound source types, with each clip representing 5 seconds. Visibility is represented by two numbers: the first indicates the number of sound sources visible for over 4 seconds within the 5-second clip, and the second denotes the number of sound sources visible for less than 4 seconds. IID Binaural acc denotes the accuracy of IID predictions respected by the actual left-right labels.}
\resizebox{0.9\linewidth}{!}{%
\begin{tabular}{clccccccc}
\toprule 
& \multirow{2}{*}{Dataset} & \multirow{2}{*}{Split} & \multicolumn{2}{c}{Size} & \multirow{2}{*}{Visibility ( \% )}  &\multicolumn{2}{c}{\small{Motion Type}} &  \multirow{2}{*}{IID binaural acc ( \% )} \\
\cmidrule(lr){4-5} \cmidrule(lr){7-8} 
 & & & \small{Clips} & \small{Duration} &  & \small{Camera} & \small{Sound sources} \\
\midrule
\multirow{8}{*}{\rotatebox[origin=c]{90}{\shortstack[c]{In the wild dataset}}} &
\multirow{2}{*}{\makecell[l]{YT-\\stereo}} & Raw & 13,000k & 8.0k hrs & -- & \multirow{2}{*}{\shortstack{Mainly\\rotation}} & -- \\
 & &Train& 14.6k & 20hrs & - &  & Unknown & Unknown \\

\cdashlinelr{2-9} 

& \multirow{4}{*}{\makecell[l]{YT-\\Stereo-iPhone}} &Raw & 95k & 80hrs & -- & -- & --  \\
& &Train & 14.6k & 20hrs & 20/50 & \multirow{2}{*}{\shortstack{Mainly \\rotation}} & Unknown & Unknown \\
&  &Val & 0.1k & 0.2hrs & 10/60 & & Moving & 57.5 \\
&  &Test & 0.1k & 0.2hrs & 10/60 & & Moving &  \\

\cdashlinelr{2-9} 
&  \makecell[l]{L/R Binaural\cite{chen2022sound}} & Raw & 1.8k & 3hrs & - & \XSolidBrush & \XSolidBrush &75.4 \\


\midrule

\multirow{4}{*}{\rotatebox[origin=c]{90}{\makecell[c]{Real-world lab-made}}} &
 \makecell[l]{Stereo-\\Fountain} &Raw & 1.4k & 2hrs & 10/30 &Mainly Stationary  & \XSolidBrush &97.0 \\ 
 \cdashlinelr{2-9} 
 
& \makecell[l]{Stereo-\\Music} &Raw & 3.6k & 5hrs & 10/30  & \XSolidBrush  & \XSolidBrush  &70.0\\
\cdashlinelr{2-9} 

& \makecell[l]{Binaural-\\Fountain} &Raw & 1.4k & 2hrs & 10/30  &Mainly Stationary  & \XSolidBrush &98.0  \\
\cdashlinelr{2-9} 

& \makecell[l]{RWAVS~\cite{liang23avnerf}}& Raw & 2.7k & 3.8hrs & Unknown  & Rotation\&Translation & \XSolidBrush  &71.9 \\
\midrule
& Simulated Dataset  & Raw & 21.6k & 30hrs & -  & Rotation\&Translation & Rendered &97.4 \\
\bottomrule
\end{tabular}
}
\label{tab:dataset}
\end{table*}

To get the motion direction of sound sources, we use interaural intensity difference (IID) cues as a pseudo label to predict the moving direction of the sound source following \cite{chen2022sound,chen2023sound}. We follow to estimate whether a certain segment of audio is on the left or right by IID as a pseudo label, which is based on which side sound is louder than the other: $d = \text{sign}(\log \left| \frac{A_{L}}{A_{R}} \right|)$ where $|A|$ is the magnitude of the spectrogram $A$.

To obtain a subset of the videos where the sound source positions are relatively easy for labelers to label, we take a way to filter the dataset for stable sound sources. Subsequently, we undertake curation through the following steps: 1) We segment the videos into 5-second clips every 3 seconds. 2) Estimating horizontal left/right movement using the Superglue model. 3) We calculate the Interaural Intensity Difference (IID) cues of the videos. 4) Due to the large number of videos, for initial filtering, we compute an IID score: every 1 second, we calculate the IID for 2 seconds, arranging the absolute product of IID changes in descending order. 5) We discard videos where the camera rotation direction differs from the IID change direction. 6) We filter out videos with camera rotation angles less than 15 degrees, and the remaining videos are sampled concerning rotation angles to ensure a relatively uniform distribution.

In \cref{fig:distribution}, we show the angle distribution of the YT-Stereo-iPhone validation set.


\paragraph{Stereo-Fountain} One of the authors positioned themselves at various fixed locations near a fountain and recorded using an iPhone 13 Pro. Subsequently, one of the authors added annotations. We split the data for train/test/val randomly.

The data collection steps were as follows: 1) One author selected 15 angles out of 360 degrees around the fountain, using an iPhone and a laser pointer (aligned visually to the fountain) to calculate the angle between the fountain and the phone; 2) The recording device was placed at each corresponding position and recorded for approximately 10 minutes.

\paragraph{Binaural-Fountain} One of the authors positioned themselves at various fixed locations near a fountain and recorded using an iPhone 13 Pro and a binaural microphone (Sennheiser AMBEO Smart Headset). Subsequently, annotations were added by one of the authors. We split the data for train/test/val randomly.

The data collection steps were as follows: 1) One author selected 15 angles out of 360 degrees around the fountain, using an iPhone and a laser pointer (aligned visually to the fountain) to calculate the angle between the fountain and the phone; 2) The author was seated at each corresponding position and recorded for approximately 10 minutes. 


\paragraph{Stereo-Music} One of the authors positioned the iPhone at a static place in the same room, and one mobile phone was playing music. Subsequently, annotations were added by one of the authors. We split data for train/test/val by mixing all the data and randomly splitting.


\paragraph{RWAVS~\cite{liang23avnerf}} This is the Real-World Audio-Visual Scene (RWAVS) dataset described in \cite{liang23avnerf}. The authors recorded data in both indoor and outdoor environments to represent daily settings. They used a 3Dio Free Space XLR binaural microphone for high-quality stereo audio, a TASCAM DR-60DMKII for recording and storing audio, and a GoPro Max for capturing videos. The devices were mounted together and moved randomly around the environment, unlike the ReplayNVAS~\cite{chen2023novel} dataset, which has a constant environment and recording viewpoint. Data collection for each scene ranged from 10 to 25 minutes.

\begin{figure*}[t]
    \centering
\includegraphics[width=1.0\linewidth]{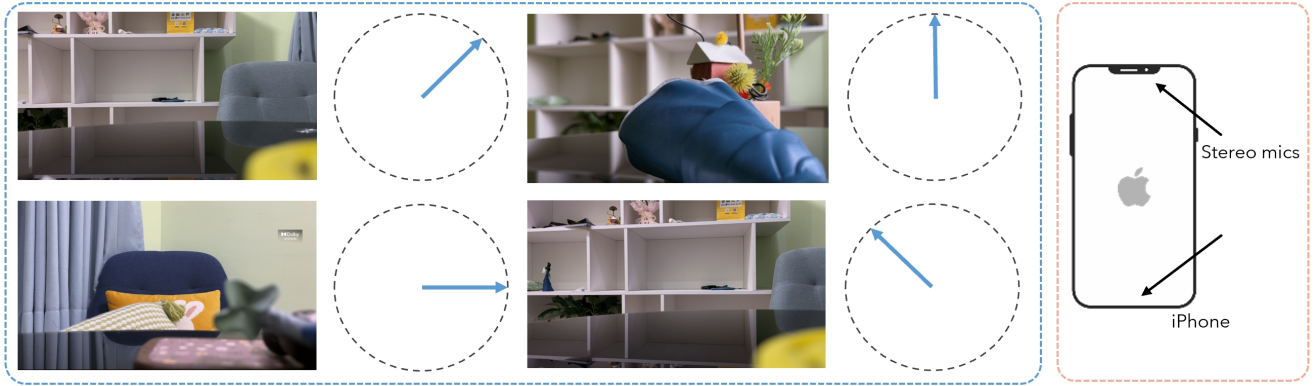}
    \caption{Examples from the {\em Stereo-Music} subset. We record sound using a stereo microphone in a scene containing music.}
\end{figure*}

\begin{figure*}[t]
    \centering
\includegraphics[width=1.0\linewidth]{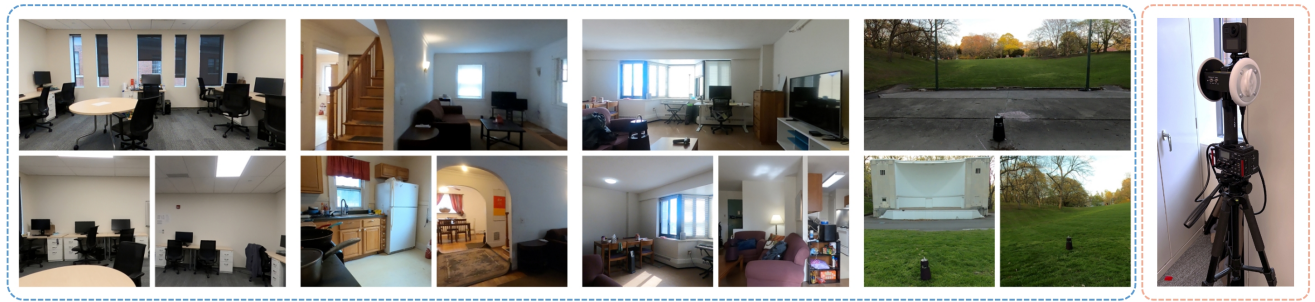}
    \caption{The examples from the RWAVS dataset were recorded using binaural microphones in various scenarios where the sound source was a loudspeaker. Above, we present images from the original paper~\cite{liang23avnerf}.}
\end{figure*}

\begin{table}[ht]
    \centering
        \caption{Summary of classification accuracy for RWAVS and Stereo-Music datasets.}
        \resizebox{\linewidth}{!}{%
    \begin{tabular}{lcc}
        \toprule
        Dataset & Classification Task & Accuracy (\%) \\
        \midrule
        RWAVS~\cite{liang23avnerf} & supervised 4-classification & 50 \\
        Stereo-Music & supervised 4-classification & 28 \\
        \bottomrule
    \end{tabular}
    }

\end{table}

\paragraph{Simulated dataset} Due to Soundspace 2.0 not supporting material and moving sound currently, we use sound sources at different positions as the moving sound. 


We follow ~\cite{chen2023sound} to create the simulated dataset. Binaural audio is obtained by convolving binaural RIRs with mono audio samples from LibriSpeech~\cite{panayotov2015librispeech}. 


Our dataset comprises 50,000 audio-visual pairs generated from 200,000 viewpoints. The audio was rendered with an average reverberation of RT60 = 0.4s. For training, validation, and testing, we divided our data into 81/9/10 scenes, respectively. We utilized approximately 30 hours of synthesized data to ensure fairness in our results.

\subsection{Label the dataset}
Annotations were added by one of the authors and another participant. This is the questionnaire used to label data. \\


You will need to label relatively clear audio events whether in the scene or out of the scene. Please use the best headphones you can have. We provide some examples to help you label the other data. \\

\begin{enumerate}[label=\arabic*)]
    \item Please provide the quality of the video's audio for scoring.
    \item Provide the video, with a 5-second audio clip, where are the audio sources located, totaling 16 classes. Or indicate if it is unknown. You only need to annotate the ones that you can distinguish where the sound is. As 90 denotes the left, -90 denotes the right. 0 denotes in front of you. 180/-180 denotes the back of you.
    \item What are the categories of the audio sources? (If repeated), please list all of them. Existing categories include car/male speech/female speech/speech/sea/animals
    \item What is the direction of movement of the audio sources? If consistent with question 3, no need to fill it.
    \item If the category cannot be indicated by numbers, there's no need to label it. If the quality is too low, just skip it.
\end{enumerate}

\subsection{Additional scenes}
To explore the relationships among various recording devices, we employ the identical procedure on a second dataset labeled YT-stereo-iPhone, also sourced from YouTube using keywords ranging from iPhone 12 Pro to iPhone 15 Pro, specifically targeting devices capable of recording spatial audio from both the bottom and top microphones. Besides, we recorded about 2 hours of labeled audio using an iPhone 13 Pro around a fountain as a supplement denoted as Stereo-Fountain and around a music player in one room, which we call {\em Stereo-Music}, and 2 hours of labeled audio using Sennheiser AMBEO Smart Headset to record binaural audio, which we call {\em Binaural-Fountain}.

\paragraph{Music data out of distribution.} For the RWAVS dataset~\cite{liang23avnerf}, we partitioned each part of the 11 scenes randomly into 80\%/10\%/10\% splits for training, testing, and validation sets, respectively. When applying a supervised approach to the Stereo-Music dataset and RWAVS Dataset, we observe that training the model solely on each dataset results in limited success. Specifically, the accuracy only reaches 50\% for the supervised 4-classification task on the RWAVS Dataset and it reaches only 28\% on the Stereo-Music dataset. This outcome highlights the current model's inadequacy when dealing with indoor music data.

\section{Experiment Implementations}\label{appendixA.2}
\subsection{Estimate camera motion} For a five-second video, we sampled five frames per second and calculated the rotation matrix and translation vector between each frame. Additionally, we computed the rotation matrix and translation vector between different images at intervals of 3 and 6 frames. During dataset cleaning, we followed a specific procedure. For any two time points, we accumulated the rotations calculated at intervals of 3 frames, proportionally adjusting if there were gaps. The same approach was applied to translations.


\subsection{Training details} For training Ours-full, the $\lambda_1$ is set as 0.9 and $\lambda_2$ is set as 1. For training Ours-R\&T, the $\lambda_1$ is set as 0.9, and $\lambda_2$ is set as 0. All models are trained using a learning rate of 0.0001 with the AdamW optimizer. The training schedule followed a cosine annealing schedule and early stopping. Training one model takes about 2 hours. Due to computational limitations, we train once for each number, using the same seed for every experiment.


%

\end{document}